\documentclass[runningheads]{llncs}
\usepackage[T1]{fontenc}
\usepackage{lscape}
\usepackage{graphicx}
\usepackage{amsmath,amssymb,amsfonts}
\usepackage{algorithmic}
\usepackage{graphicx}
\usepackage{textcomp}
\usepackage{xcolor}
\usepackage{float}
\restylefloat{table}
\usepackage{lscape}
 \usepackage[numbers]{natbib}
\usepackage{graphicx}
\usepackage{caption}
\usepackage{subcaption}
\usepackage{booktabs,makecell}
\usepackage{multirow}
\usepackage{pdflscape}
\usepackage{caption}
\usepackage{hyperref}
\usepackage{rotating}
\usepackage[export]{adjustbox}
\usepackage{comment}
\usepackage{amsmath}
\usepackage{xcolor}
\usepackage{tikz}
\usetikzlibrary{shapes,arrows}
\usepackage{placeins}
\usepackage{graphics}
\usepackage{xcolor}
\usepackage{mathrsfs}
\usepackage{blindtext}
\usepackage{hyperref}

\usepackage{algorithm}
\usepackage{algorithmic}

\usepackage{epstopdf}
\def\BibTeX{{\rm B\kern-.05em{\sc i\kern-.025em b}\kern-.08em
    T\kern-.1667em\lower.7ex\hbox{E}\kern-.125emX}}
\begin{document}
\title{Uncertainty-Aware Ensemble Deep Randomized Neural Networks for Classification}
\author{M. Sajid$^{1}$ \and A. Quadir$^{1}$ \and A. Rahaman$^{1}$ \and P. N. Suganthan$^{2}$ \and M. Tanveer$^{1}$}
\authorrunning{M. Sajid et al.}
%
\institute{Indian Institute of Technology Indore, Simrol, Indore, India\
\email{\{phd2101241003,mscphd2207141002,phd2401141001,mtanveer\}@iiti.ac.in}
\and
Qatar University, Qatar\
\email{p.n.suganthan@qu.edu.qa}
}

\maketitle              
\begin{abstract}
The current state-of-the-art (SOTA) deep randomized neural networks, such as deep Random Vector Functional Link (dRVFL) and ensemble deep RVFL (edRVFL), treat all training samples uniformly, which limits their robustness and effectiveness when applied to real-world datasets containing noise and outliers. Furthermore, the propagation of contaminated features across hidden layers negatively influences the decision-making capability of these models. To overcome these limitations, we propose intuitionistic fuzzy dRVFL (IF-dRVFL) and intuitionistic fuzzy edRVFL (IF-edRVFL) frameworks that enhance model robustness. The proposed models unify intuitionistic fuzzy theory to exploit sample neighborhood information in the kernel space by jointly considering membership and non-membership degrees for each sample. Membership degrees are computed based on the distance of samples from their respective class centroids, while non-membership degrees quantify sample heterogeneity within local neighborhoods. These measures are employed to assign adaptive weights to training samples, enabling effective discrimination among clean, noisy, and outlier data points. Extensive experiments conducted on UCI and KEEL benchmark datasets, with and without the presence of Gaussian noise, demonstrate the superiority of the proposed IF-dRVFL and IF-edRVFL models over existing SOTA fuzzy and non-fuzzy approaches. The source code is available at  \url{https://github.com/mtanveer1/IF-edRVFL}.

\keywords{Random vector functional link (RVFL) \and Deep RVFL, Ensemble deep RVFL \and Uncertainty-aware \and Robustness, Intuitionistic fuzzy.}
\end{abstract}
\section{Introduction}
Artificial Neural Networks (ANNs) have achieved remarkable success across a wide range of machine learning tasks; however, their practical deployment is often hindered by slow convergence, sensitivity to learning rates, and susceptibility to local minima \cite{malik2022random, 10416391}. To overcome these limitations, Randomized Neural Networks (RdNNs), such as the Random Vector Functional Link (RVFL) network, have been proposed \cite{pao1994learning}. In RVFL, the weights connecting the input layer to the hidden layer are randomly generated and fixed, while direct links from the input to the output layer act as an inherent regularization mechanism \cite{zhang2016comprehensive}. This architecture enables efficient training through closed-form solutions and offers competitive generalization performance.

Despite these advantages, the standard RVFL model employs a shallow architecture with a single hidden layer, which restricts its ability to capture complex nonlinear relationships present in real-world data and often results in unstable classification performance. To address these shortcomings, two major extensions of RVFL have been explored: ensemble learning \cite{qiu2018ensemble} and deep learning \cite{shi2021random, 10552388}. Ensemble RVFL models improve stability and robustness by aggregating multiple base learners, while deep RVFL (dRVFL) and ensemble deep RVFL (edRVFL) introduce multiple hidden layers to enhance representational power. In particular, edRVFL exploits implicit ensemble learning by treating each hidden layer as an individual RVFL classifier, leading to improved generalization and stability \cite{shi2021random}. Recently, edRVFL-based variants have been successfully employed across a wide range of applications, including enhancing interpretability through fuzzy inference systems \cite{10552388}, time-series forecasting \cite{10880477}, Alzheimer’s disease diagnosis \cite{10839045}, and image recognition systems \cite{zhao2025ensemble}.

Although (e)dRVFL-based models offer faster training and fewer parameters compared to conventional deep networks, they remain vulnerable to noise and outliers commonly encountered in real-world datasets. The presence of noisy samples can significantly degrade their learning process and predictive performance. This limitation motivates the development of robust deep RVFL-based models capable of effectively handling noisy and outlier-contaminated data.
\section{Motivation and Contributions}
\subsection{Motivation}
The susceptibility of dRVFL and edRVFL models to noise and outliers arises from two fundamental issues. First, noisy or corrupted features present in training samples propagate through successive hidden layers, resulting in a mixture of pure and impure features that contaminates higher-level representations and adversely affects decision-making. Second, existing dRVFL-based models assign uniform importance to all training samples, disregarding the inherent differences between clean, noisy, and outlier instances. This uniform treatment leads to suboptimal generalization, particularly in challenging environments with significant noise or anomalous data points.

Fuzzy theory has been widely employed in machine learning to alleviate the adverse effects of noise and outliers by assigning adaptive importance to data samples \cite{rezvani2019intuitionistic}. In particular, intuitionistic fuzzy (IF) theory \cite{ha2013support} extends conventional fuzzy sets by jointly considering membership and non-membership degrees, offering a more expressive framework for modeling uncertainty and data heterogeneity. By quantifying both the degree of belongingness and non-belongingness of a sample to a class, IF theory provides a principled mechanism to distinguish reliable samples from noisy or outlier instances. This capability makes intuitionistic fuzzy theory a natural and effective choice for enhancing the robustness of deep RVFL-based models \cite{ganaie2024graph}.

\begin{figure}
\begin{minipage}{.48\linewidth}
\centering
\subfloat[dRVFL]{\includegraphics[scale=0.22]{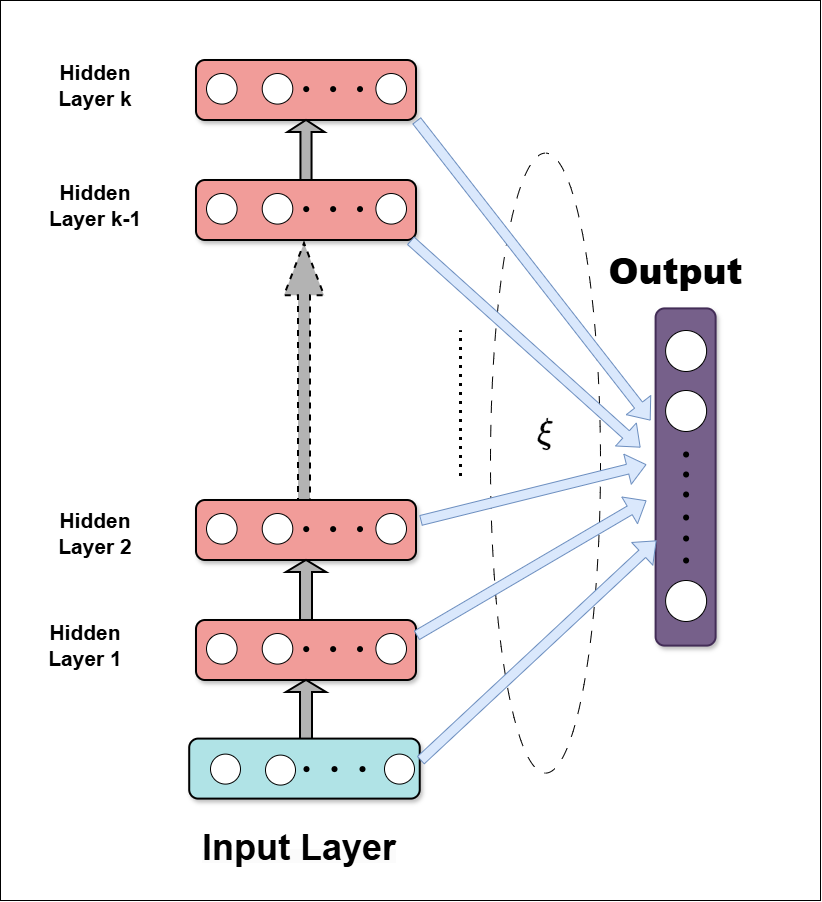}}
\end{minipage}
\begin{minipage}{.48\linewidth}
\centering
\subfloat[edRVFL]{\includegraphics[scale=0.195]{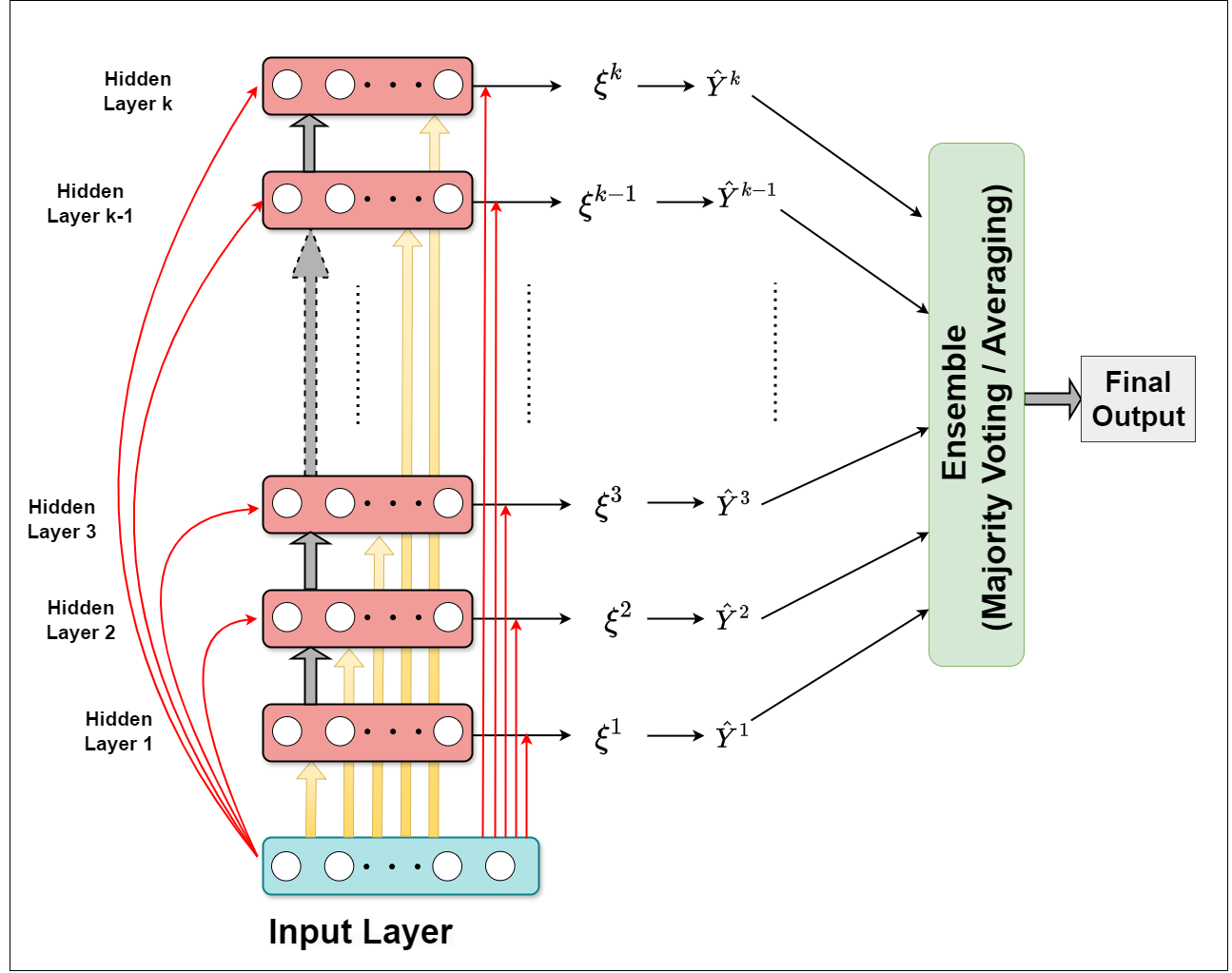}}
\end{minipage}
\caption{Framework of dRVFL and edRVFL.}
\label{fig:edRVFL}
\end{figure}

\subsection{Contributions}
The main contributions of this paper are as follows:
\begin{itemize}
\item \textbf{\textit{Novel robust models:}} We propose two novel models, namely IF-based dRVFL (IF-dRVFL) and edRVFL (IF-edRVFL).
\item  \textbf{\textit{Adaptive sample weighting:}} In the proposed models, each sample is assigned an IF score in the kernel space, derived from its membership value (based on the distance to the corresponding class centroid) and non-membership value (capturing neighborhood heterogeneity). These IF scores are used to assign adaptive weights to training samples, enabling effective discrimination among clean, noisy, and outlier data.
\item \textit{\textbf{Robust deep architecture:}} Unlike conventional deep models where impure features propagate through hidden layers, the proposed IF-dRVFL framework consists of stacked robust hidden layers, each trained using IF-weighted samples, thereby mitigating the adverse impact of noise propagation.
\item \textit{\textbf{Efficient ensemble learning:}} The proposed IF-edRVFL model achieves ensemble learning implicitly by training a single robust deep RVFL network, avoiding the computational overhead of training multiple independent models while simultaneously benefiting from deep and ensemble learning principles.
\end{itemize}

\noindent\textbf{\textit{Note:}} The Figures in \ref{fig:edRVFL} represent the architectures of the dRVFL and edRVFL models, and the supplementary Section S.I briefly goes through the mathematical formulation of dRVFL and edRVFL.

\section{Proposed Models}
This section begins with defining notations and then presents the detailed mathematical formulation of the proposed IF-(e)dRVFL models, followed by the description of the IF weighting scheme. Let $s_u$ denote the IF score assigned to the training sample $x_u$, and let $S=\mathrm{diag}(s_1,s_2,\ldots,s_N)$ represent the corresponding diagonal IF weight matrix for the training set $\mathcal{X}$ (see Section~\ref{Intuitionistic Fuzzy Membership (IFM) Scheme}).

\subsection{Notations:} Let the training dataset be $\mathcal{X} =\{(x_u, y_u) | u \in \{1,2, \ldots, N\}\}$, where $y_u \in \mathbb{R}^{1 \times c}$ represents the target vector of $x_u \in \mathbb{R}^{1 \times m}$, with $N$ number of total training samples. $m$ is the number of attributes, and the number of classes is denoted by $c$. $(\cdot)^{T}$ represent the transpose operator. $X=[x_1^T,x_2^T, \hdots, x_N^T]^T$ and  $Y=[y_1^T,y_2^T, \hdots, y_N^T]^T$ are matrices of input and output samples, respectively.\\

\subsection{Proposed IF-dRVFL Network}
In the standard dRVFL framework, all training samples are treated uniformly, irrespective of their quality or reliability. However, real-world datasets often contain noise and outliers, which can adversely affect the learning capability and generalization performance of dRVFL models. To mitigate the influence of such contaminated samples, we propose the IF-dRVFL network. In the proposed IF-dRVFL model, the fuzzy membership value of a sample is determined based on its proximity to the centroid of its corresponding class, reflecting the degree of belongingness. Conversely, the non-membership value is computed by incorporating neighborhood information, capturing the degree of non-belongingness of the sample. By jointly considering membership and non-membership information, the IF-dRVFL framework assigns adaptive weights to training samples, thereby enhancing robustness against noise and outliers.

Let $E^{(1)}$ represent the first hidden layer matrix, acquired through the projection of the input matrix using randomly initialized weights, followed by the activation function $\psi$ as:
\begin{align}
\label{eq:First hidden layer}
    E^{(1)} = \psi(X\omega^{(1)}),
\end{align}
here, $\omega^{(1)} \in \mathbb{R}^{m \times h}$ represents the weights for the first hidden layer, initialized randomly from a uniform distribution of $[-1, 1]$ and $h$ represents the number of hidden nodes. \\
The deeper hidden layers ($g > 1$) are defined as:
\begin{align}
    E^{(g)} = \psi(E^{(g-1)}\omega^{(g)}),
\end{align}
where $\omega^{(g)}$ represents the randomly generated weights for the $g^{th}$ hidden layer. The enhanced feature set is defined by concatenating all the original and hidden layer features:
\begin{align}
    E^* = [E^{(k)}, E^{(k-1)}, \dots , E^{(1)}, X].
\end{align}
Now, we propose the optimization problem of IF-dRVFL as:
\begin{align}
\label{eq:12}
      & \underset{\xi}{min} \hspace{0.2cm} \frac{\mathcal{C}}{2} \|S\eta\|^2 + \frac{1}{2} \|\xi\|^2 \nonumber \\
     & s.t.  \hspace{0.2cm} E^*\xi - Y = \eta, 
\end{align}
where $\mathcal{C}$ is a regularization parameter and $\xi$ is the output layer weights (unknown) and needs to be calculated. The Lagrangian corresponding to the problem \eqref{eq:12} is formulated as:
\begin{align}
    L = \frac{\mathcal{C}}{2} \|S(E^*\xi - Y)\|^2 + \frac{1}{2} \|\xi\|^2. 
\end{align}
Now, we differentiate \( L \) with respect to \( \xi \) and then equate it to zero, we obtain:
\begin{align}
\label{eq:IF-dRVFL-11}
    \frac{\partial{L}}{\partial{\xi}} = \xi + \mathcal{C}(SE^*)^T(S(E^*\xi - Y)) = 0.
\end{align}
Solving \eqref{eq:IF-dRVFL-11}, we get\\
\begin{align}
    \xi & = \left (  (SE^*)^T(SE^*) + \frac{1}{\mathcal{C}}I \right )^{-1}(SE^*)^TSY \nonumber\\
    & = \left (  E{^*}^TS^2E^* + \frac{1}{\mathcal{C}}I \right )^{-1}E{^*}^TS^2Y.
\end{align}
where $I$ is the identity matrix of the appropriate dimension.\\
Substituting the value $\xi=(SE^*)^T\eta$ \cite{malik2022random}, we get
\begin{align}
\label{eq:simplify}
    (SE^*)\eta + \mathcal{C}(SE^*)^T(S(E^*(SE^*)^T\eta -Y)) = 0.
\end{align}
After simplifying \eqref{eq:simplify}, we get:
\begin{align}
    \eta = \left (  SE^*E^{*T}S + \frac{1}{\mathcal{C}}I \right )^{-1}SY \nonumber 
    = S^{-1}\left (S^{2}E^*E^{*T} + \frac{1}{\mathcal{C}}I \right )^{-1}S^{2}Y.
\end{align}
Finally,
\begin{align}
    \xi=(SE^*)^T\eta = E^{*T}\left (S^{2}E^*E^{*T} + \frac{1}{\mathcal{C}}I \right )^{-1}S^{2}Y.
\end{align}
Thus, we can provide the optimal solution of \eqref{eq:12} as follows:
\begin{equation}
\xi=\left\{\begin{array}{ll}\left (  E{^*}^TS^2E^* + \frac{1}{\mathcal{C}}I \right )^{-1}E{^*}^TS^2Y, & (m+kh) \leq N,\vspace{3mm} \\ 
E^{*T}\left (S^{2}E^*E^{*T} + \frac{1}{\mathcal{C}}I \right )^{-1}S^{2}Y, & N<(m+kh). \end{array}\right.
\end{equation}
The flowchart of the proposed IF-dRVFL is given in Fig. \ref{fig:flowchart} (a).
\subsection{Proposed IF-edRVFL Network}
In the proposed IF-edRVFL, each layer is treated as a base model, thus unifying the ensemble learning principle and intuitionistic fuzzy scheme into deep RVFL. This facilitates the creation of multiple diverse and robust base models (RVFLs) within a single deep framework. The final outcome of IF-edRVFL is determined through either averaging or a majority voting scheme.  Within IF-edRVFL, every base model (hidden layer) receives inputs comprising the original features and the randomized features computed by the preceding layer. Furthermore, the optimization problem of each base model possesses the IF score weights. Thus, each base models behave as a robust base model and enriches the learning process. 

The first hidden layer matrix $E^{(1)}$ is defined as in \eqref{eq:First hidden layer}
and the deeper hidden layers' ($g > 1$) output is defined as:
\begin{align}
    E^{(g)}=\psi([E^{(g-1)}, X]\omega^{(g)}), ~~g=1,2,\hdots,k.
\end{align}
where $\omega^{(1)}$ and $\omega^{(g)}$ ($g>1$) represent the randomly generated weights for the first and $g^{th}$ ($g>1$) hidden layers, respectively. The feature set corresponding to the first base model and the $g^{th}$ base model is defined as follows: 
$F^{(1)}=[E^{(1)}, X]$ and $F^{(g)}=[E^{(g)}, E^{(g-1)}, X]$, respectively.

\begin{figure}
\begin{minipage}{.48\linewidth}
\centering
\subfloat[IF-dRVFL]{\includegraphics[scale=0.25]{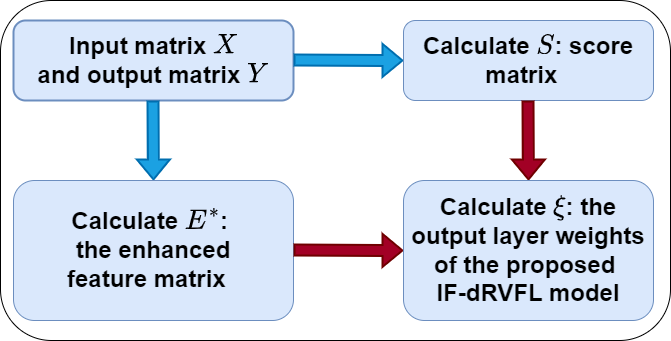}}
\end{minipage}
\begin{minipage}{.48\linewidth}
\centering
\subfloat[IF-edRVFL]{\includegraphics[scale=0.22]{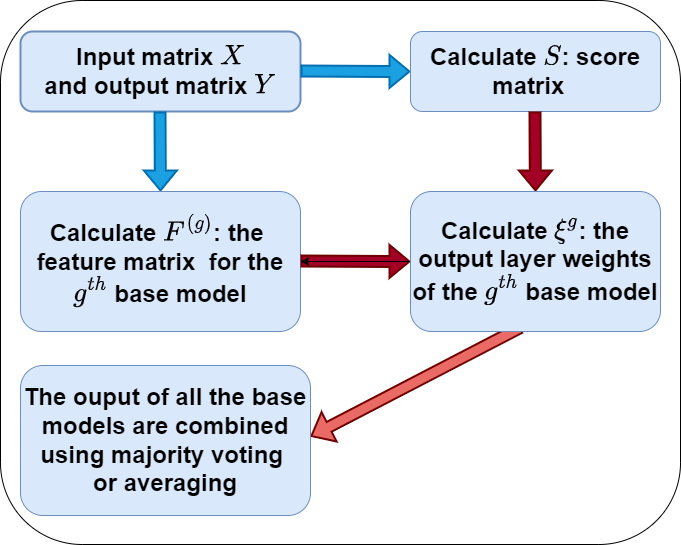}}
\end{minipage}
\caption{Flowchart of the proposed models.}
\label{fig:flowchart}
\end{figure}

The proposed optimization problem corresponding to the $g^{th}$ base model is given as follows:
\begin{align}
\label{eq:21}
      & \underset{\xi^{g}}{min} \hspace{0.2cm} \frac{\mathcal{C}}{2} \|S\eta^g\|^2 + \frac{1}{2} \|\xi^{g}\|^2 \nonumber \\
     & s.t.  \hspace{0.2cm} F^{(g)}\xi^{g} - Y = \eta^g, 
\end{align}
where $\xi^{g}$ is the output layer weight of the $g^{th}$ base model. The solution of \eqref{eq:21} follows the same procedure as solved \eqref{eq:12} of the proposed IF-dRVFL model. The output of all the base models is combined using majority voting to make the final decision of the proposed IF-edRVFL model. The flowchart of the proposed IF-edRVFL is given in Fig. \ref{fig:flowchart} (b).
\subsection{Intuitionistic Fuzzy Scheme}
\label{Intuitionistic Fuzzy Membership (IFM) Scheme}
The intuitionistic fuzzy approach \cite{ha2013support} assigns IF score to each sample based on three parameters: the membership degree $\nu (0 \leq \nu \leq 1)$, the non-membership degree $\mu (0 \leq \mu \leq 1)$, and the degree of hesitation $\pi = 1 - \nu - \mu$. These parameters are then combined using a score function to assign an IF score value to each sample, aiding in the assessment of noise and outlier presence within the dataset. 

\textbf{(i) Membership function:} The membership function is a way to quantify how much a specific sample belongs to its corresponding class. It assigns a numerical value or degree of belongingness that indicates the sample's level of similarity or fit with the characteristics of that class. This is calculated as the distance between the training sample and the class centroid in the high-dimensional feature space. This is helpful in identifying outliers or instances that do not conform to the typical traits of a specific class. Mathematically, we define the degree of membership for each training sample as follows:
    \begin{equation}
    \mu(x_{u}) = 
            \begin{cases}
                  1 - \frac{\|\phi(x_{u}) - C^+\|}{r^+ + \hspace{0.2cm} \gamma}, & y_{u} = +1 ~(\text{positive class}),\\
                   1 - \frac{\|\phi(x_{u}) - C^-\|}{r^- + \hspace{0.2cm} \gamma}, & y_{u} = -1 ~(\text{negative class}),
            \end{cases}
    \end{equation}
    where $\phi$ represents the projection mapping function, $\gamma$ is a non-negative parameter, and $r^+ (r^-)$ denotes the radius of $+1$ $(-1)$ class given by:
    \begin{align}
    r^{+}=\underset{y_{u}= +1}\max \Vert \ \phi(x_{u})-C^{+}\Vert, \hspace{0.2cm} \text{and} \hspace{0.2cm} r^{-}=\underset{y_{u}= -1}\max \Vert \ \phi(x_{u})-C^{-}\Vert,
    \end{align}
    where $C^+ (C^-)$ denote the centers of the $+1 (-1)$ classes, respectively and are defined as:
    \begin{align}
    C^+ = \frac{1}{m_{1}} \sum _{y_{u}= +1} \phi(x_{u}), \hspace{0.2cm} \text{and} \hspace{0.2cm} C^- = \frac{1}{m_{2}} \sum _{y_{u}= -1} \phi(x_{u}),
\end{align}
here $m_{1} (m_{2})$ represent the number of samples in the $+1 (-1)$ class, respectively.\\
\textbf{(ii) Non-membership function:} The non-membership function provides a measure of dissimilarity or mismatch between the sample and the characteristics of the class it is a part of. The non-membership function for each training sample quantifies the ratio of dissimilar samples to the total number of samples in its neighborhood. The non-membership value aids in identifying noisy samples. The non-membership function is defined as follows: 
\begin{equation}
    \nu(x_{u}) = (1 - \mu (x_{u})) \alpha (x_{u}),
\end{equation}
where the value $\alpha(x_u)$ is calculated as:
\begin{equation} 
\alpha (x_{u})=\frac{|\lbrace x_{j}|\Vert  \phi(x_{u})- \phi(x_{j})\Vert \leq \beta,\,y_{j}\ne y_{u}\rbrace |}{|\lbrace x_{j}|\Vert  \phi(x_{u})- \phi(x_{j})\Vert \leq \beta \rbrace |}.
\end{equation}
Here, $\beta$ is a non-negative parameter that can be adjusted.\\
\textbf{(iii) The score mapping:} Now, we define the IF score mapping by combining each sample's membership and non-membership values. This resulting IF score acts as a unified measure to determine if a sample is pure, noisy, or an outlier. The IF score values for the $x_u$ is assigned as follows:
\begin{equation}
s_{u}=\left\lbrace \begin{array}{lr} \mu (x_{u}), & \nu (x_{u})=0,\\ 0, & \mu (x_{u})\leq \nu (x_{u}), \\ \frac{1-\nu (x_{u})}{2-\mu (x_{u})-\nu (x_{u})}, & \text{others}. \end{array}\right. 
\end{equation}
Finally, the score matrix $S$ for the dataset $\mathcal{X}$ is defined as: $S=diag\{s_u \hspace{0.1cm} | \hspace{0.1cm} u=1,2, \ldots, N\}$.
The kernel technique is explored in supplementary Section S.II.
\begin{table*}[]
\centering
\caption{Average accuracy and standard deviation with SOTA non-fuzzy RdNNs on 13 KEEL datasets.}
\label{tab:AVg_KEEL_ACC}
\resizebox{\textwidth}{!}{%
\begin{tabular}{lccccccccc} \hline \vspace{-3mm}\\  \textbf{\textbf{Metric} $\downarrow$ $\mid$ \textbf{Model} $\rightarrow$} &
  RVFL \cite{pao1994learning} &
  RVFLwoDL \cite{huang2006extreme} &
  BLS \cite{chen2017broad} &
  H-ELM \cite{tang2015extreme} &
  dRVFL \cite{shi2021random} &
  edRVFL \cite{shi2021random} &
  edEGEFRVFL \cite{9893913} &
  IF-dRVFL (ours) &
  IF-edRVFL (ours)  \vspace{1mm}\\ \hline \vspace{-3mm} \\
  \textbf{Average Accuracy} &
  81.87 &
  81.51 &
  82.55 &
  81.65 &
  83.29 &
  83.04 &
  83.07 &
  \textbf{84.32} &
  \textbf{84.39}
  \vspace{1mm}\\ \hline \vspace{-3mm} \\
  \textbf{Average Standard Deviation}         & 5.1   & 5.13  & 4.46  & 5.07  & 4.99  & 5.19  & 5.14  & \textbf{6.22} & \textbf{5.64}
\vspace{1mm}\\ \hline \vspace{-3mm} \\
\end{tabular}%
}
\end{table*}
\section{Experimental Results}
To assess the effectiveness of the proposed IF-dRVFL and IF-edRVFL models, we perform a comparative analysis with baseline models using benchmark datasets from the UCI \cite{dua2017uci} and KEEL \cite{derrac2015keel} repository.

\subsection{Hyperparameter Selection and Experimental Setup}
In the proposed IF-dRVFL and IF-edRVFL models, intuitionistic fuzzy weights are generated by transforming the samples into a higher-dimensional space through the use of a kernel function. We used the Gaussian kernel and is given by $K(x_i,x_j) = e^{\frac{-1}{2\sigma^2}\|x_i - x_j\|^2}.$ Gaussian kernel parameter $\sigma$ is selected from the range $\{2^{-5}, 2^{-4}\ldots, 2^{5}\}$. The regularization parameters for each model are selected from the set $\mathcal{C} = \{10^{-5}, 10^{-4}, \ldots, 10^{-5}\}$. Following the methodology outlined in \cite{tang2015extreme}, the selection of the number of hidden layer nodes ``$h$'' in the RVFLwoDL, RVFL, and H-ELM follows the range $h = 100:100:2000$. Following \cite{9893913}, the graph regularization parameter for the edEGERVFL model is set equal to $\mathcal{C}$. Following \cite{shi2021random}, a two-stage tuning approach was employed for the dRVFL, edRVFL, edEGERVFL, edRVFL-FIS-C, IF-dRVFL, and IF-edRVFL models. In the first stage, the parameters are fine-tuned to obtain the optimal hidden nodes $h^*$ within the range of $[256, 512, 1024]$, and the regularization parameter $\mathcal{C}^*$ is determined while keeping the number of hidden layers fixed at two. For IF-dRVFL and IF-edRVFL, $\sigma=1$ is fixed, and $\sigma$ is fine-tuned in the second stage. During the second stage, hidden layers within the range of $1:1:10$ are tuned, along with other parameters, in the vicinity of $g^*$ and $\mathcal{C}^*$.
Furthermore, for the proposed edRVFL-FIS models, the best number of fuzzy nodes $K^{*}$ in the first stage is tuned within the range of $5:10:45$, and in the subsequent stage, they are fine-tuned in the vicinity of $K^{*}$, specifically within the range of $K^{*}-4:1:K^{*}+5$. Feature groups for BLS and F-BLS are selected from the range $N_2 = 1:2:21$, while the selection of feature nodes within each feature group is made from the range $N_1 = 5:5:50$. The NF-BLS model selects the number of fuzzy groups from the range $N_{fg} = 1:2:21$ and the number of fuzzy nodes in each fuzzy group from the range $N_{fn} = 5:5:50$. The number of enhancement nodes for BLS, F-BLS and F-BLS is chosen from the range $N_3 = 5:5:50$. 
\\
\textbf{Setup:} The experimental setup is discussed in the Supplementary Section S.III.

\subsection{Models in Comparison} The performance of the proposed IF-dRVFL and IF-edRVFL models is compared against 11 baselines with two kinds of models: Firstly, 7 SOTA non-fuzzy RdNNs, including RVFL \cite{pao1994learning}, RVFL without direct link (RVFLwoDL) -also known as  extreme learning machine (ELM) \cite{huang2006extreme}, broad learning system (BLS) \cite{chen2017broad}, hierarchical ELM (H-ELM) \cite{tang2015extreme}, dRVFL \cite{shi2021random}, edRVFL \cite{shi2021random}, ensemble deep extended graph embedded RVFL
(edEGERVFL) \cite{9893913}. Secondly, 4 SOTA fuzzy-based baselines such as neuro-fuzzy BLS (NF-BLS) \cite{feng2018fuzzy}, Fuzzy BLS \cite{sajid2024intuitionistic}, IF twin support vector machine (IF-TSVM) \cite{rezvani2019intuitionistic} and edRVFL based on fuzzy inference system (edRVFL-FIS-C) \cite{10552388}.
\subsection{Experimental Results and Statistical Analysis on Real-World Datasets Against SOTA Non-Fuzzy RdNNs}
In this subsection, we evaluate the proposed IF-dRVFL and IF-edRVFL models against state-of-the-art (SOTA) non-fuzzy randomized deep neural network (RdNN) baselines on $13$ KEEL benchmark datasets spanning diverse domains and sample sizes.

\textit{\textbf{Accuracy (ACC):}} 
The average classification accuracy (ACC), standard deviation (Std.), and corresponding ranks are summarized in Table~\ref{tab:AVg_KEEL_ACC}, while the detailed dataset-wise results are reported in Supplementary Tables S.1--S.3. 
It is evident that the proposed IF-edRVFL and IF-dRVFL models achieve the highest and second-highest average ACC values of $84.39\%$ and $84.32\%$, respectively. 
In comparison, the baseline RVFL, RVFLwoDL, BLS, H-ELM, dRVFL, edRVFL, NF-BLS, and edEGEFRVFL models attain average ACC values of $81.87\%$, $80.51\%$, $82.55\%$, $81.65\%$, $8329\%$, $83.04\%$, and $81.07\%$, respectively. 
These results clearly demonstrate the superior and more robust performance of the proposed intuitionistic fuzzy models across most datasets.


\textit{\textbf{Statistical rank:}} Sometimes, the average accuracy metric of a model can be influenced by outstanding performance in a single dataset, which could compensate for weaker results across various datasets, potentially resulting in a biased measure. Therefore, we employ a ranking method to assess the effectiveness of the compared models. In this approach, each classifier is assigned a rank, with the model demonstrating superior performance receiving a lower rank and the model exhibiting inferior performance receiving a higher rank. To evaluate $q$ models across $P$ datasets, let $r_j^i$ represent the rank of the $j^{th}$ model on the $i^{th}$ dataset. $\mathscr{R}_j = \frac{1}{P}\sum_{i=1}^P r_j^i$ is the average rank of the $j^{th}$ model. Supplementary Table S.3, we note that the average rank of proposed IF-dRVFL and IF-edRVFL models along with the RVFL, RVFLwoDL, BLS, H-ELM, dRVFL, edRVFL, and edEGEFRVFL models is $3.12$, $3.23$, $5.81$, $6.88$, $3.96$, $6.62$, $4.81$, $5.35$, and $5.23$, respectively. The proposed IF-dRVFL and IF-edRVFL achieve the lowest and second lowest average rank (best performance) among all compared models, showcasing superior generalization ability.
\\ \textit{\textbf{Friedman test:}} Now, we perform the Friedman test \cite{demvsar2006statistical} to determine if there are statistically significant differences among the models. Under the null hypothesis, it is assumed that all models have equal average ranks, indicating equivalent levels of performance. The Friedman statistic follows the chi-squared distribution $(\chi_F^2)$ with $(q - 1)$ degrees of freedom (d.o.f), and its computation involves: $\chi_F^2 = \frac{12P}{q(q+1)}\left[ \sum_j \mathscr{R}_j^2 - \frac{q(q+1)^2}{4} \right]$. The $F_F$ statistic is computed as: $F_F = \frac{(P-1)\chi_F^2}{P(q-1) - \chi_F^2}$, where the $F$-distribution possesses degrees of freedom $(q - 1)$ and $(P - 1) \times (q - 1)$. In our case, we have $q=9$ and $P=13$, thus the obtained values are $\chi_F^2 = 25.78$ and $F_F = 3.96$. The critical value $F_F(8, 96) = 2.04$ at a $5\%$ level of significance. The null hypothesis is rejected as $3.96 > 2.04$. Thus, there exists a statistically significant difference among the models being compared. \\
\textit{\textbf{Nemenyi post hoc test:}} Next, we utilize the Nemenyi post hoc test to examine the pairwise differences between the models. The critical difference $(C.D.)$ value is calculated as $C.D. = q_{\alpha}\sqrt{\frac{q(q+1)}{6P}}$. For $q_{\alpha} = 3.10$ for $9$ models at a significance level of $5\%$, we get $C.D. = 3.33$. The average rank differences between the proposed IF-dRVFL and IF-edRVFL models and the baselines RVFLwoDL and H-ELM confirm their statistical superiority. Furthermore, given the lowest average ranks achieved by the IF-dRVFL and IF-edRVFL models, we assert that these proposed models surpass the baseline models in overall performance.

\subsection{Experimental Results and Statistical Analysis on Real-World Datasets Against SOTA Fuzzy Baselines}
The results in Table \ref{tab:Avg_UCI_Accuracy} clearly demonstrate the superior performance of the proposed IF-dRVFL and IF-edRVFL models over SOTA fuzzy baselines on 12 UCI datasets. IF-edRVFL achieves the highest average accuracy of $83.18\%$, followed by IF-dRVFL with $82.48\%$, consistently outperforming NF-BLS, F-BLS, IF-TSVM, and edRVFL-FIS-C. This improvement highlights the effectiveness of incorporating intuitionistic fuzzy modeling within deep randomized neural networks for handling uncertainty and data ambiguity.

From the statistical analysis (refer Supplementary Section S.IV), IF-edRVFL and IF-dRVFL obtain the lowest average ranks of $2.46$ and $2.75$, respectively, indicating superior generalization and stability across datasets. The Friedman test ($\chi_F^2=18.14$, $F_F=4.77$) rejects the null hypothesis at the $5\%$ significance level, confirming statistically significant differences among the models. Furthermore, the Nemenyi post hoc test ($C.D.=2.17$) verifies that the proposed models significantly outperform IF-TSVM. Overall, the proposed intuitionistic fuzzy deep RVFL frameworks deliver higher accuracy, improved robustness, and statistically superior performance compared to existing fuzzy baselines.

\subsection{Robustness Evaluation of Proposed Models on Datasets with Gaussian Noise}
\begin{table}[]
\centering
\caption{Average accuracy, standard deviation and rank comparison with SOTA fuzzy baselines on 12 UCI datasets.}
\label{tab:Avg_UCI_Accuracy}
\resizebox{9cm}{!}{%
\begin{tabular}{lcccccc} \hline \vspace{-3mm}\\  \textbf{\textbf{Metric} $\downarrow$ $\mid$ \textbf{Model} $\rightarrow$}
 &
  NF-BLS \cite{feng2018fuzzy} &
  F-BLS \cite{sajid2024intuitionistic} &
  IF-TSVM \cite{rezvani2019intuitionistic} &
  edRVFL-FIS-C \cite{10552388} &
  IF-dRVFL &
  IF-edRVFL  \vspace{1mm}\\ \hline \vspace{-3mm} \\
  \textbf{Average Accuracy}                      & 78.58 & 82    & 73.19 & 80.19 & \textbf{82.48} & \textbf{83.18}
\vspace{1mm}\\ \hline \vspace{-3mm} \\
\textbf{Average Standard Dev.}                      & 12.61 & 8.98  & 25.47 & 12.16 & \textbf{11.76} & \textbf{12.09}
\vspace{1mm}\\ \hline \vspace{-3mm} \\
\textbf{Average Rank}                      & 3.42 & 3.42 & 5.46 & 3.46 & \textbf{2.75} & \textbf{2.46}
\vspace{1mm}\\ \hline \vspace{-3mm} \\
\end{tabular}%
}
\end{table}
\begin{figure}
    \centering
    \includegraphics[width=1.1\linewidth]{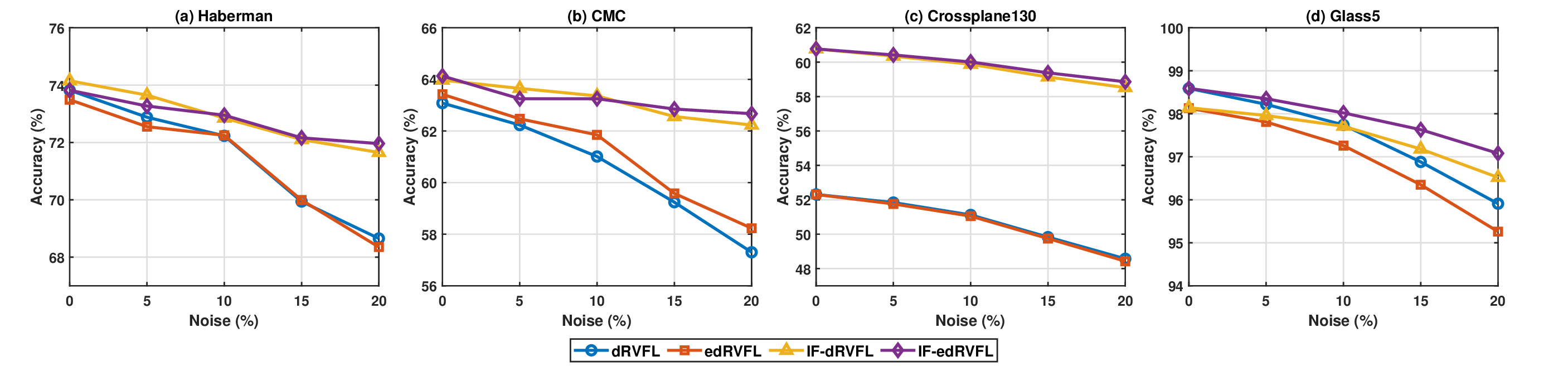}
    \caption{Robustness: Effect of different noise levels on the performance of dRVFL, edRVFL, IF-dRVFL (proposed) and IF-edRVFL (proposed) models.}
    \label{fig:noise}
\end{figure}
To further assess the robustness of the proposed IF-dRVFL and IF-edRVFL models, we compare them against the baseline dRVFL and edRVFL under different levels of Gaussian noise. For a balanced evaluation, four representative datasets are considered: Haberman and CMC, where the proposed models are initially outperformed by the baseline methods; Crossplane130, where the proposed models achieve superior performance; and Glass5, where the best baseline and proposed models exhibit comparable accuracies (see Supplementary Table S.1). Gaussian noise levels of 5\%, 10\%, 15\%, and 20\% are introduced, and the results are presented in Fig.~\ref{fig:noise}.

For the Haberman and CMC datasets, the baseline models continue to achieve slightly higher accuracies than the proposed methods across all noise levels. However, a closer examination reveals that IF-dRVFL and IF-edRVFL experience considerably smaller performance degradation as the noise level increases. In contrast, dRVFL and edRVFL exhibit a much sharper decline in accuracy, particularly beyond 10\% noise. This indicates that the intuitionistic fuzzy framework provides greater resilience to data corruption and preserves predictive performance more effectively under noisy conditions.

A similar robustness advantage is observed on the Crossplane130 dataset, where the proposed models already outperform the baselines in the noise-free setting. As the noise level increases, IF-dRVFL and IF-edRVFL maintain substantially higher accuracies, causing the performance gap over the baseline methods to widen further. For the Glass5 dataset, where the best baseline and proposed models achieve comparable performance at 0\% noise, the intuitionistic fuzzy variants exhibit a noticeably slower degradation rate and consistently retain higher accuracies under increasing noise levels.

Overall, the results demonstrate that incorporating intuitionistic fuzzy modeling into deep RVFL architectures significantly improves robustness to noise. Whether the proposed models initially outperform, match, or underperform the baseline methods, they consistently exhibit greater stability and a slower loss of predictive performance as the noise level increases.

\subsection{Sensitivity Analyses}
We conduct the sensitivity analyses to delve deeper into the behaviour of the models: (i) by investigating the impact of intuitionistic fuzzy Gaussian kernel parameter $\sigma$, (ii) and by exploring the impact of the number of hidden layers $L$.
\begin{figure}
\begin{minipage}{.242\linewidth}
\centering
\subfloat[haberman\\ (IF-dRVFL)]{\includegraphics[scale=0.165]{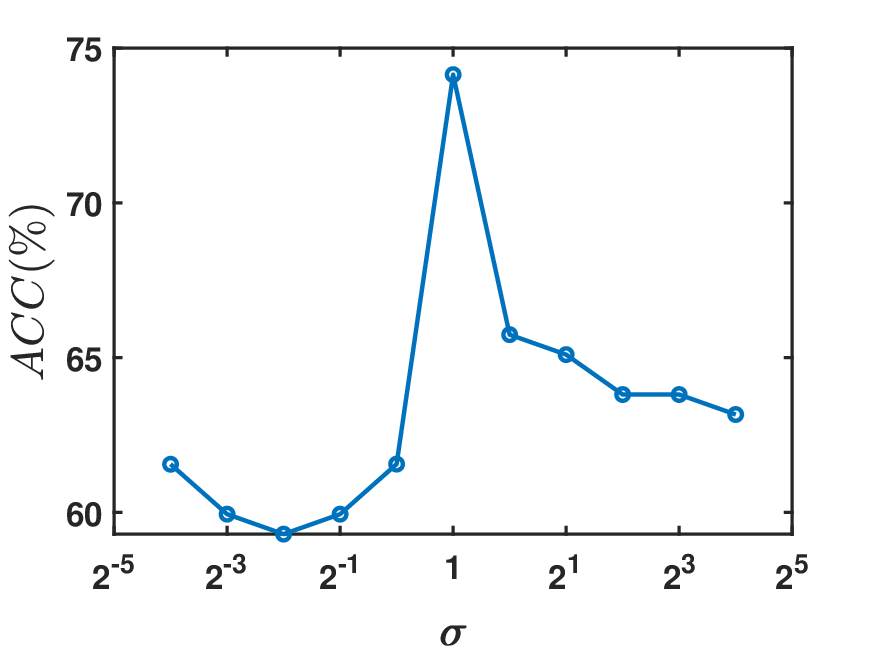}}
\end{minipage}
\begin{minipage}{.242\linewidth}
\centering
\subfloat[titanic\\ (IF-dRVFL)]{\includegraphics[scale=0.165]{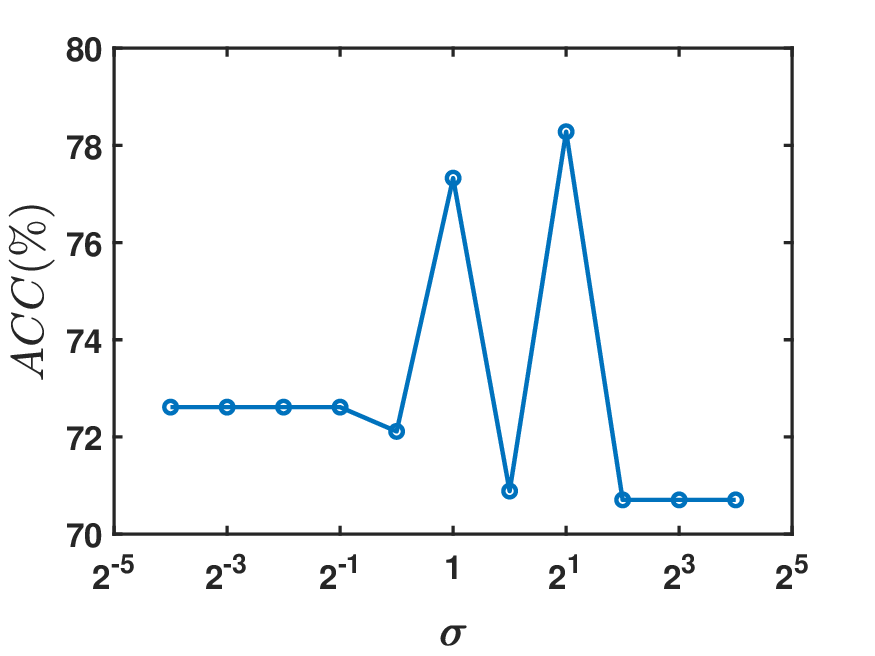}}
\end{minipage}
\begin{minipage}{.242\linewidth}
\centering
\subfloat[haberman\\ (IF-edRVFL)]{\includegraphics[scale=0.165]{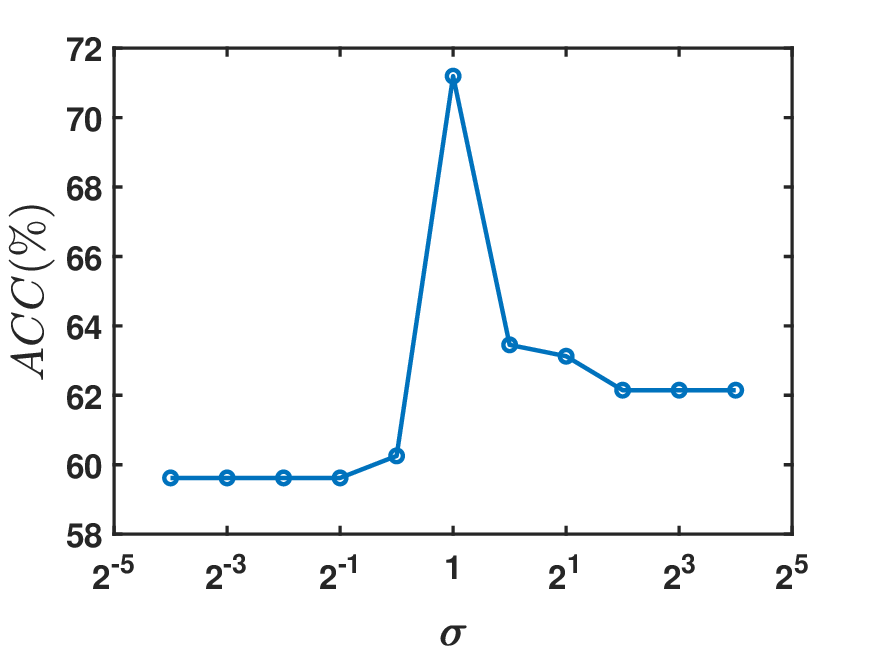}}
\end{minipage}
\begin{minipage}{.242\linewidth}
\centering
\subfloat[titanic\\ (IF-edRVFL)]{\includegraphics[scale=0.165]{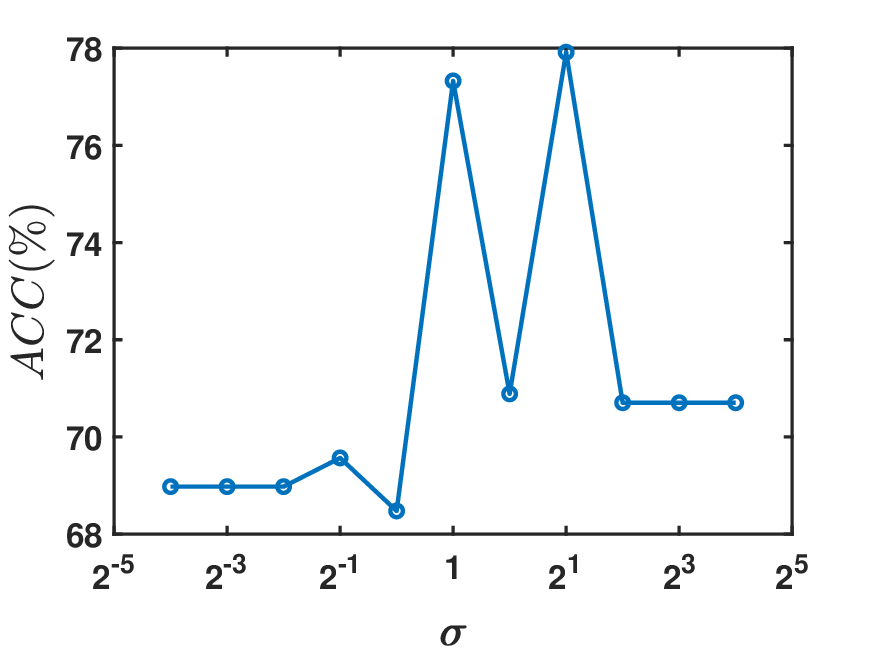}}
\end{minipage}
\caption{Effect of parameter $\sigma$ on the performance of the proposed IF-dRVFL and IF-edRVFL models.}
\label{effect of sigma parameter}
\end{figure}
\begin{figure}
\begin{minipage}{.242\linewidth}
\centering
\subfloat[haberman\\ (IF-dRVFL)]{\includegraphics[scale=0.165]{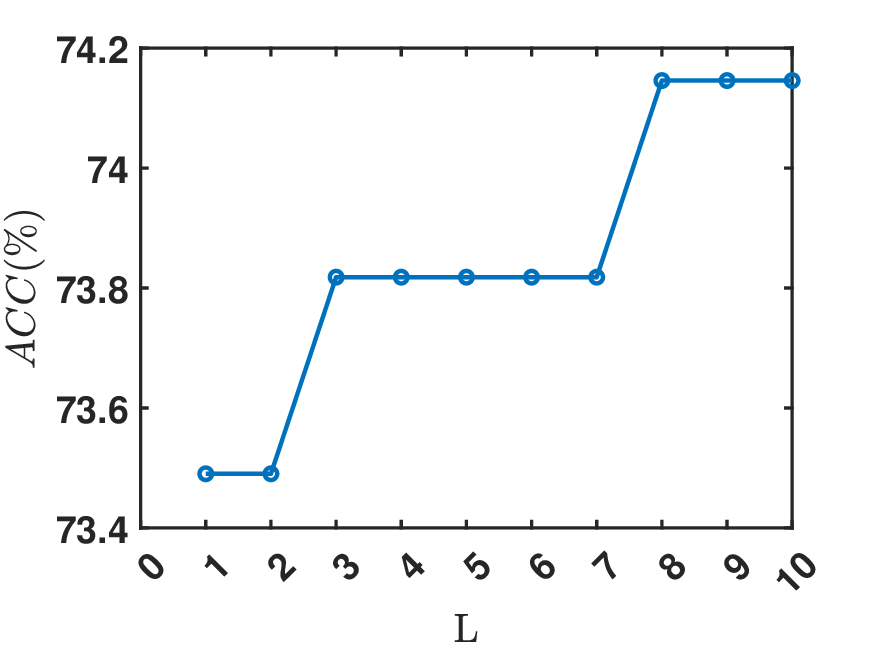}}
\end{minipage}
\begin{minipage}{.242\linewidth}
\centering
\subfloat[titanic\\ (IF-dRVFL)]{\includegraphics[scale=0.165]{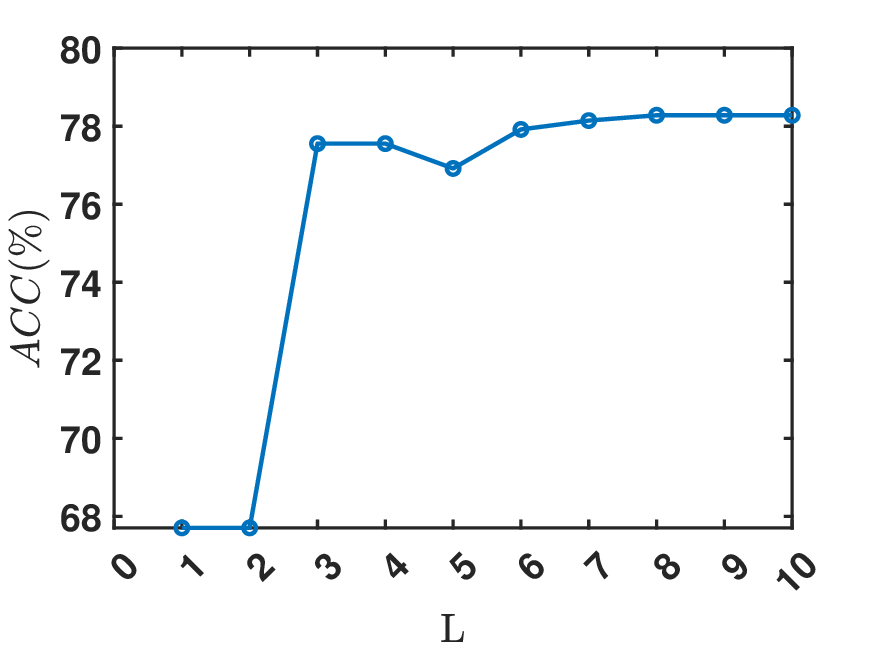}}
\end{minipage}
\begin{minipage}{.242\linewidth}
\centering
\subfloat[haberman\\ (IF-edRVFL)]{\includegraphics[scale=0.165]{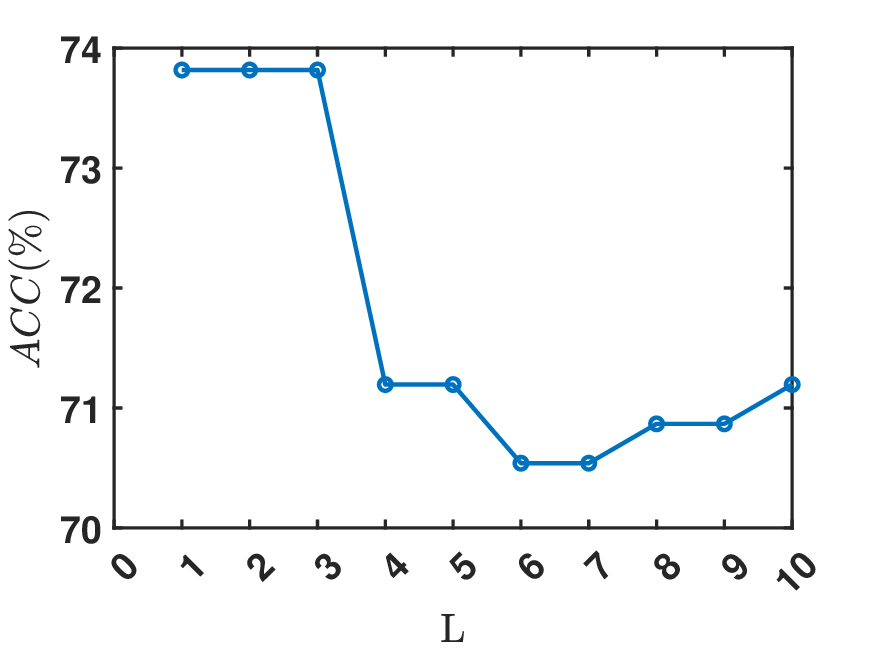}}
\end{minipage}
\begin{minipage}{.242\linewidth}
\centering
\subfloat[titanic\\ (IF-edRVFL)]{\includegraphics[scale=0.165]{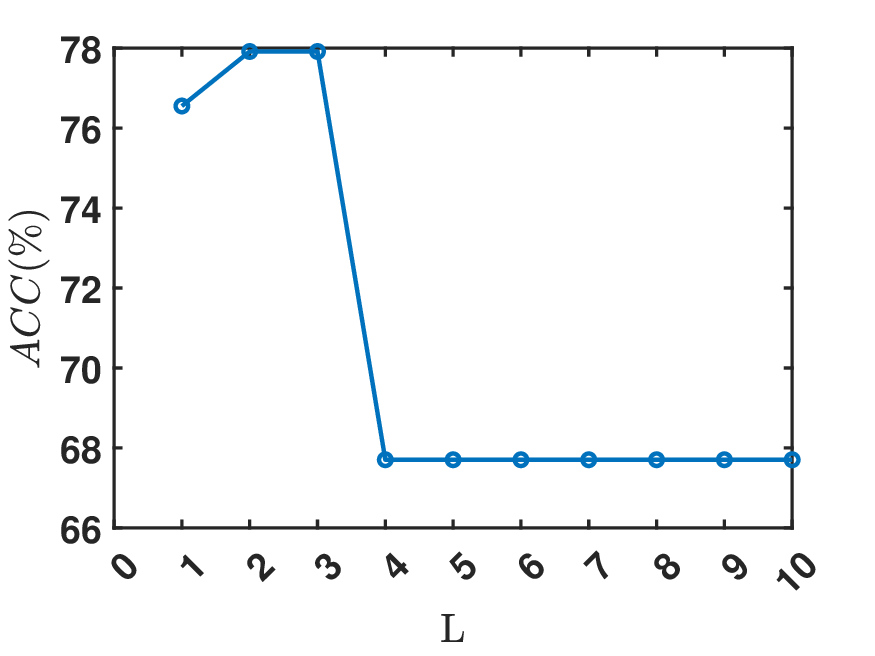}}
\end{minipage}
\caption{Effect of parameter $L$ on the proposed IF-(e)dRVFL models.}
\label{effect of L parameter}
\end{figure}

\textit{\textbf{(1) Influence of intuitionistic fuzzy Gaussian kernel parameter $\sigma$:}} The impact of $\sigma$ on the performance of the proposed models is shown in Fig. \ref{effect of sigma parameter}. Specifically, Fig. \ref{effect of sigma parameter} (a) and (b) pertain to the IF-dRVFL models on the haberman and titanic datasets, respectively, while Fig. \ref{effect of sigma parameter} (c) and (d) focus on the IF-edRVFL models on the same datasets. We note that the performance of all proposed models improves as $\sigma$ increases from $2^{-5}$. Optimal performance is often achieved when $\sigma$ is set to $2^0=1$ or $2^1=2$. Subsequently, as $\sigma$ further increases, performance begins to diminish. Therefore, we recommend using $\sigma=1$ or $2$ for optimal results, although fine-tuning may be necessary depending on the dataset's characteristics.

\textit{\textbf{(2) Influence of the number of hidden layers $L$:}} The impact of the hyperparameter $L$ is shown in Fig. \ref{effect of L parameter}. Our analysis reveals the following: (a) For the proposed IF-dRVFL model, performance consistently improves with an increase in the number of hidden layers until reaching a plateau. Optimal performance is typically achieved with higher values of $L$, such as $8$ or larger. (b) In the case of the proposed IF-edRVFL model, we observe that performance peaks at $L=3$ and then gradually declines as $L$ increases further. Therefore, to achieve the best performance from the IF-edRVFL model, we recommend using $L=3$. However, we recommend fine-tuning the hyperparameters to achieve the best performance for the proposed models for specific tasks.
\section{Conclusion}
In this paper, we proposed the IF-dRVFL and IF-edRVFL models to address the sensitivity issues observed in the baseline dRVFL and edRVFL models. By integrating fuzzy concepts into the deep and ensemble frameworks of these models, we enhance their robustness against noise and outliers present in the dataset. Our experimentation involved testing the proposed IF-dRVFL and IF-edRVFL models on benchmark datasets from UCI and KEEL repository, comparing them against $11$ state-of-the-art models within the SOTA RdNNs and fuzzy domains. We evaluated the robustness of our models under different conditions, including scenarios with and without contaminated Gaussian noise in the dataset's features. Our experimental results and statistical analyses highlight the superior performance of the proposed IF-dRVFL and IF-edRVFL models, ranking them as the top-performing and second-best models, respectively.

\begin{credits}
\subsubsection{\ackname} M. Sajid acknowledges the Council of Scientific and Industrial Research (CSIR), New Delhi, for providing fellowship under grants 09/1022(13847)/2022-EMR-I.

\subsubsection{\discintname}
The authors have no competing interests to declare that are
relevant to the content of this article.
\end{credits}
%
%
%
\bibliographystyle{splncs04}
\bibliography{refs.bib}
\end{document}


\title{{\small{Supplementary}} \\
Uncertainty-Aware Ensemble Deep Randomized Neural Networks for Classification}
\author{M. Sajid$^{1}$ \and A. Quadir$^{1}$ \and A. Rahaman$^{1}$ \and P. N. Suganthan$^{2}$ \and M. Tanveer$^{1}$}
\authorrunning{M. Sajid et al.}
%
\institute{Indian Institute of Technology Indore, Simrol, Indore, India\
\email{\{phd2101241003,mscphd2207141002,phd2401141001,mtanveer\}@iiti.ac.in}
\and
Qatar University, Qatar\
\email{p.n.suganthan@qu.edu.qa}
}
\maketitle              
%

\section{Related Work}
In this section, we first define some notations and briefly go through the mathematical formulation of dRVFL and edRVFL.
\subsection{Notations}
Let the training dataset be $\mathcal{X} =\{(x_u, y_u) | u \in \{1,2, \ldots, N\}\}$, where $y_u \in \mathbb{R}^{1 \times c}$ represents the target vector of $x_u \in \mathbb{R}^{1 \times m}$, with $N$ number of total training samples. $m$ is the number of attributes, and the number of classes is denoted by $c$. $(\cdot)^{T}$ represent the transpose operator. $X=[x_1^T,x_2^T, \hdots, x_N^T]^T$ and  $Y=[y_1^T,y_2^T, \hdots, y_N^T]^T$ is the collection of all input and output samples, respectively.
\subsection{Deep random vector functional link (dRVFL) network \cite{shi2021random}}
Within the dRVFL architecture, multiple hidden layers are vertically stacked. These hidden layers' weights are initially generated randomly and remain unchanged throughout the training phase. Afterward, the features from all hidden layers, combined with the original features, are fused together, resulting in an augmented feature set. The output layer's weights are then computed using an analytical method. Let there be $k$ number of hidden layers, then the first hidden layer's output of dRVFL is defined as:
\begin{align}
    E^{(1)} = \psi(X\omega^{(1)}),
\end{align}
and the higher hidden layers' ($g > 1$) output is defined as:
\begin{align}
    E^{(g)} = \psi(E^{(g-1)}\omega^{(g)}), ~~g=1,2,\hdots,k.
\end{align}
Here, $h$ is the number of hidden nodes in the hidden layer. $\omega^{(1)}$ and $\omega^{(g)}$ are the randomly initialized weights for the first and $g^{th}$ hidden layers, respectively. $\psi(\cdot)$ is the activation function. The enhanced feature matrix is defined as:
\begin{align}
    E^* = [E^{(k)}, E^{(k-1)}, \dots , E^{(1)}, X].
\end{align}
The dRVFL's output is defined as:
\begin{align}
    \hat{Y}=E^*\xi,
\end{align}
where $\xi \in \mathbb{R}^{(kh+m) \times c}$ denotes weights of the output layer and is calculated as:
\begin{equation}
\label{eq:solution_RVFL}
\xi=E{^{*}}^{\dagger}\hat{Y},
\end{equation}
where $E{^{*}}^{\dagger}$ denotes the pseudo-inverse of $E^{*}$. We can employ the least-square method as well to calculate $\xi$.
\subsection{Ensemble dRVFL (edRVFL) \cite{shi2021random}}
The Ensemble deep RVFL (edRVFL) model was conceived by merging implicit ensemble learning principles into the dRVFL architecture, effectively combining the strengths of both ensemble and deep learning frameworks. Within edRVFL, every base model (hidden layer) receives inputs comprising both the original features and the randomized features computed by the preceding layer. This integration enriches the learning process. The output from the first hidden layer in edRVFL is given as follows:
\begin{align}
    E^{(1)}=\psi(X\omega^{(1)}),
\end{align}
and the higher hidden layers' ($g > 1$) output is defined as:
\begin{align}
    E^{(g)}=\psi([E^{(g-1)}, X]\omega^{(g)}), ~~g=1,2,\hdots,k,
\end{align}
where where $\omega^{(1)}$ and $\omega^{(g)}$ ($g>1$) represent the randomly generated weights for the first and $g^{th}$ ($g>1$) hidden layers, respectively. Each hidden layer in the edRVFL comprises a base model (RVFL). The output of the $1^{st}$ base model (generated from the first hidden layer) is defined as:
\begin{align}
    \hat{Y^{1}}=[E^{(1)}, X]\xi^{1},
\end{align}
where $\xi^{1}$ are the output layer weights for the first base model. The $g^{th}$ base model's output is defined as follows:
\begin{align}
    \hat{Y^{g}}=[E^{(g)}, E^{(g-1)}, X]\xi^{g}.
\end{align}
Here, $\xi^{g}$ are the output layer weights for the $g^{th}$ base model and calculated using \eqref{eq:solution_RVFL}. Finally, the decision is taken by consolidating the outcomes from all base models (hidden layers) through averaging or majority voting methods. 
\section{Kernel Technique Used for Finding Radius in Assigning Intuitionistic Fuzzy Score}
The Kernel technique is explored here.
\begin{theorem}
Let the kernel function is $K(x_{u}, x_j)$. Then, the inner product distance is given by:\\
$\|\phi(x_u) -\phi(x_j)\|=\sqrt{K(x_{u}, x_u) - 2K(x_{u}, x_j) + K(x_{j}, x_j)}$.
\end{theorem}
$Proof.$
\begin{align}
         &\|\phi(x_u) -\phi(x_j)\|=\sqrt{(\phi(x_u) -\phi(x_j)).(\phi(x_u) -\phi(x_j))} \nonumber \\
            & = \sqrt{(\phi(x_u).\phi(x_u)) - (\phi(x_u).\phi(x_j)) + (\phi(x_j).\phi(x_j))}  \nonumber \\
            & = \sqrt{K(x_{u}, x_u) - 2K(x_{u}, x_j) + K(x_{j}, x_j)}. \nonumber
\end{align}

\begin{theorem}
    The Euclidean distance between the samples and the corresponding class center is represented by: \\
    {{
    $\|\phi(x_u) - C^+\|=\\ \sqrt{K(x_{u}, x_u) - \frac{2}{m_1}\sum_{y_j = +1}K(x_{u}, x_j) + \frac{1}{m_1^2}\sum_{y_u = +1}\sum_{y_j = +1}K(x_{u}, x_j)}$, \\
    $\|\phi(x_u) - C^-\| =\\ \sqrt{K(x_{u}, x_u) - \frac{2}{m_2}\sum_{y_j = -1}K(x_{u}, x_j) + \frac{1}{m_2^2}\sum_{y_u = -1}\sum_{y_j = -1}K(x_{u}, x_j)}$,}} \\
where $m_1$ and $m_2$ represent the number of samples of $+1$ and $-1$ classes, respectively.    
\end{theorem}
$Proof.$
{{
\begin{align}
    &\|\phi(x_u) - C^+\|\\ &=\sqrt{(\phi(x_u) - C^+).(\phi(x_u) - C^+)}  \nonumber \\
    & = \sqrt{(\phi(x_u).\phi(x_u)) + (C^+.C^+) -  2(\phi(x_u).C^+)}  \nonumber \\
    & = \sqrt{K(x_{u}, x_u) +  \left (\frac{1}{m_1}\sum_{y_u = +1}\phi(x_u)\right). \left (\frac{1}{m_1}\sum_{y_u = +1}\phi(x_u)\right)  -  2\phi(x_u) \left (\frac{1}{m_1}\sum_{y_j = +1}\phi(x_u)\right)} \nonumber \\
    & = \sqrt{K(x_{u}, x_u) +  \frac{1}{m_1^2}\sum_{y_u = +1}\sum_{y_j = +1}K(x_{u}, x_j)  -  \frac{2}{m_1}\sum_{y_j = +1}K(x_{u}, x_j)}. \nonumber
\end{align}}}
Similarly, $ \|\phi(x_u) - C^-\|$
can be calculated.
\section{Experimental Setup}
\subsection{Experimental Software} The experimental setup consists of a PC equipped with an Intel(R) Xeon(R) Gold $6226$R CPU operating at a speed of $2.90$GHz and featuring $128$ GB of RAM. This system runs on the Windows $11$ platform and the experiments are executed using MATLAB R$2023$a.
\subsection{Experimental Procedure} Grid search with a 5-fold cross-validation technique is used to optimize the model's hyperparameters. In order to accomplish this, the dataset was divided into five distinct, non-overlapping subsets, or ``folds". One subset was set to be designated for testing, and the other four were used for training in every phase. For every fold, testing accuracy was separately calculated using different hyperparameter sets. The average testing accuracy was determined by taking the mean of these five accuracies for every combination of hyperparameters. The models' testing accuracy is determined by selecting the model with the highest average testing accuracy.

\section{Statistical Tests with Fuzzy Baselines}

\textit{\textbf{Statistical ranking:}} 
The comparative ranking results against fuzzy baseline models are summarized in Supplementary Table S.VI. The average ranks obtained by the proposed IF-edRVFL and IF-dRVFL models are $2.46$ and $2.75$, respectively, whereas the NF-BLS, F-BLS, IF-TSVM, and edRVFL-FIS-C models achieve average ranks of $3.42$, $5.46$, and $3.46$. 
The consistently lower ranks attained by the proposed models indicate superior and more stable performance across the evaluated datasets, highlighting their enhanced generalization capability.

\textit{\textbf{Friedman test:}} 
To examine the statistical significance of the observed performance differences, the Friedman test is conducted on $q=6$ classifiers evaluated over $P=12$ datasets. The resulting test statistics are $\chi_F^2 = 18.14$ and $F_F = 4.77$. 
Since the obtained $F_F$ value exceeds the critical threshold $F_F(5,55)=2.38$ at the $5\%$ significance level, the null hypothesis of equal performance among the models is rejected. This confirms the presence of statistically significant differences across the compared fuzzy learning approaches.

\textit{\textbf{Nemenyi post hoc test:}} 
Subsequently, the Nemenyi post hoc test is employed to identify pairwise performance differences among the models. For a significance level of $5\%$ and $q_{\alpha}=2.85$, the corresponding critical difference is computed as $C.D.=2.17$. 
The rank differences reveal that both IF-edRVFL and IF-dRVFL exhibit statistically significant improvements over the IF-TSVM baseline. Furthermore, given their lowest average ranks among all models, the proposed intuitionistic fuzzy deep RVFL frameworks consistently outperform existing fuzzy baselines in terms of overall accuracy, robustness, and ranking performance. 

\begin{table*}[]
\centering
\caption{Accuracy comparison with SOTA non-fuzzy baseline RdNNs on KEEL datasets.}
\label{tab:KEEL_Accuracy}
\resizebox{\textwidth}{!}{%
\begin{tabular}{lccccccccc} \hline \vspace{-3mm}\\  \textbf{\textbf{Dataset} $\downarrow$ $\mid$ \textbf{Model} $\rightarrow$}
 &
  RVFL \cite{pao1994learning} &
  RVFLwoDL \cite{huang2006extreme} &
  BLS \cite{chen2017broad} &
  H-ELM \cite{tang2015extreme} &
  dRVFL \cite{shi2021random} &
  edRVFL \cite{shi2021random} &
  edEGEFRVFL \cite{9893913} &
  IF-dRVFL (ours) &
  IF-edRVFL (ours)  \vspace{1mm}\\ \hline \vspace{-3mm} \\
abalone9-18              & 95.49 & 94.21 & 95.18 & 95.36 & 96.04 & 95.63 & 95.77 & 96.04 & 96.04 \\
bupa\_or\_liver-disorders  & 71.88 & 71.88 & 74.49 & 73.33 & 73.62 & 73.62 & 73.04 & 74.2  & 73.91 \\
cmc                      & 64.58 & 63.22 & 65.66 & 63.22 & 63.08 & 63.42 & 63.56 & 63.97 & 64.13 \\
crossplane130            & 39.23 & 38.21 & 39.23 & 39.23 & 52.31 & 52.31 & 52.31 & 60.77 & 60.77 \\
crossplane150            & 62    & 62    & 62    & 62    & 62.67 & 60    & 62    & 64.67 & 67.33 \\
ecoli-0-1-4-7\_vs\_2-3-5-6 & 97.31 & 97.31 & 97.21 & 97.31 & 97.01 & 97.31 & 95.83 & 97.02 & 97.32 \\
ecoli-0-1-4-7\_vs\_5-6     & 97.89 & 96.81 & 98.8  & 97.29 & 97.89 & 97.89 & 97.89 & 98.5  & 98.8  \\
ecoli-0-6-7\_vs\_5         & 97.27 & 97.27 & 98.18 & 95.91 & 97.27 & 96.82 & 97.27 & 98.18 & 97.27 \\
glass2                   & 92.05 & 92.05 & 92.05 & 92.05 & 92.05 & 92.05 & 92.05 & 92.05 & 92.05 \\
glass4                   & 95.77 & 95.77 & 97.07 & 95.78 & 97.65 & 98.12 & 97.65 & 97.67 & 96.26 \\
glass5                   & 96.72 & 96.72 & 98.13 & 95.79 & 98.59 & 98.13 & 97.65 & 98.14 & 98.59 \\
haberman                 & 74.15 & 74.15 & 74.15 & 74.15 & 73.82 & 73.49 & 73.82 & 74.15 & 73.82 \\
heart-stat               & 80    & 80    & 80.96 & 80    & 80.74 & 80.74 & 81.11 & 80.74 & 80.74 \vspace{1mm}\\ \hline \vspace{-3mm} \\
\textbf{Average} &
  81.87 &
  81.51 &
  82.55 &
  81.65 &
  83.29 &
  83.04 &
  83.07 &
  \textbf{84.32} &
  \textbf{84.39}
  \vspace{1mm}\\ \hline \vspace{-3mm} \\
\end{tabular}%
}
\end{table*}
\begin{table*}[]
\centering
\caption{Standard deviation comparison with SOTA non-fuzzy baseline RdNNs on KEEL datasets.}
\label{tab:KEEL_Std}
\resizebox{\textwidth}{!}{%
\begin{tabular}{lccccccccc} \hline \vspace{-3mm}\\  \textbf{\textbf{Dataset} $\downarrow$ $\mid$ \textbf{Model} $\rightarrow$}
 &
  RVFL \cite{pao1994learning} &
  RVFLwoDL \cite{huang2006extreme} &
  BLS \cite{chen2017broad} &
  H-ELM \cite{tang2015extreme} &
  dRVFL \cite{shi2021random} &
  edRVFL \cite{shi2021random} &
  edEGEFRVFL \cite{9893913} &
  IF-dRVFL (ours) &
  IF-edRVFL (ours)  \vspace{1mm}\\ \hline \vspace{-3mm} \\
abalone9-18              & 3.95  & 3.95  & 3.36  & 3.9   & 3.17  & 3.52  & 3.45  & 3.24          & 3.06          \\
bupa\_or\_liver-disorders  & 3.01  & 3.01  & 2.2   & 2.43  & 3.46  & 5.65  & 3.01  & 5.74          & 3.97          \\
cmc                      & 14.99 & 14.99 & 12.69 & 14.01 & 13.75 & 14.28 & 17.72 & 16.69         & 14.88         \\
crossplane130            & 7.4   & 7.4   & 7.4   & 7.4   & 7.98  & 7.98  & 7.98  & 21.31         & 14.03         \\
crossplane150            & 12.16 & 12.16 & 12.16 & 12.16 & 13.21 & 12.16 & 12.16 & 11.45         & 12.16         \\
ecoli-0-1-4-7\_vs\_2-3-5-6 & 1.64  & 1.64  & 1.25  & 1.64  & 1.83  & 1.64  & 2.22  & 1.06          & 0.68          \\
ecoli-0-1-4-7\_vs\_5-6     & 1.36  & 1.36  & 0.67  & 1.26  & 0.82  & 0.82  & 0.82  & 0.06          & 0.67          \\
ecoli-0-6-7\_vs\_5         & 1.9   & 1.9   & 1.02  & 1.9   & 1.02  & 2.03  & 1.02  & 1.9           & 2.96          \\
glass2                   & 2.12  & 2.12  & 2.12  & 2.12  & 2.12  & 2.12  & 2.12  & 2.12          & 2.12          \\
glass4                   & 4.23  & 4.23  & 1.27  & 3.85  & 2.35  & 1.99  & 2.35  & 2.33          & 2.66          \\
glass5                   & 1.3   & 1.3   & 1.05  & 1.95  & 1.28  & 1.95  & 1.68  & 1.95          & 2.09          \\
haberman                 & 8.06  & 8.06  & 8.06  & 8.06  & 8.52  & 8.48  & 8.52  & 8.06          & 8.52          \\
heart-stat               & 4.22  & 4.61  & 4.79  & 5.3   & 5.34  & 4.83  & 3.8   & 5             & 5.49      \vspace{1mm}\\ \hline \vspace{-3mm}    \\
\textbf{Average}         & 5.1   & 5.13  & 4.46  & 5.07  & 4.99  & 5.19  & 5.14  & \textbf{6.22} & \textbf{5.64}
\vspace{1mm}\\ \hline \vspace{-3mm} \\
\end{tabular}%
}
\end{table*}
\begin{table*}[]
\centering
\caption{Rank comparison with SOTA non-fuzzy baseline RdNNs on KEEL datasets.}
\label{tab:KEEL_Rank}
\resizebox{\textwidth}{!}{%
\begin{tabular}{lccccccccc} \hline \vspace{-3mm}\\  \textbf{\textbf{Dataset} $\downarrow$ $\mid$ \textbf{Model} $\rightarrow$}
 &
  \multicolumn{1}{l}{RVFL \cite{pao1994learning}} &
  \multicolumn{1}{l}{RVFLwoDL \cite{huang2006extreme}} &
  \multicolumn{1}{l}{BLS \cite{chen2017broad}} &
  \multicolumn{1}{l}{H-ELM \cite{tang2015extreme}} &
  \multicolumn{1}{l}{dRVFL \cite{shi2021random}} &
  \multicolumn{1}{l}{edRVFL \cite{shi2021random}} &
  \multicolumn{1}{l}{edEGEFRVFL \cite{9893913}} &
  \multicolumn{1}{l}{IF-dRVFL (ours)} &
  \multicolumn{1}{l}{IF-edRVFL (ours)} \vspace{1mm}\\ \hline \vspace{-3mm} \\
abalone9-18              & 6    & 9    & 8    & 7    & 2    & 5    & 4    & 2             & 2             \\
bupa\_or\_liver-disorders  & 8.5  & 8.5  & 1    & 6    & 4.5  & 4.5  & 7    & 2             & 3             \\
cmc                      & 2    & 7.5  & 1    & 7.5  & 9    & 6    & 5    & 4             & 3             \\
crossplane130            & 7    & 9    & 7    & 7    & 4    & 4    & 4    & 1.5           & 1.5           \\
crossplane150            & 6    & 6    & 6    & 6    & 3    & 9    & 6    & 2             & 1             \\
ecoli-0-1-4-7\_vs\_2-3-5-6 & 3.5  & 3.5  & 6    & 3.5  & 8    & 3.5  & 9    & 7             & 1             \\
ecoli-0-1-4-7\_vs\_5-6     & 5.5  & 9    & 1.5  & 8    & 5.5  & 5.5  & 5.5  & 3             & 1.5           \\
ecoli-0-6-7\_vs\_5         & 5    & 5    & 1.5  & 9    & 5    & 8    & 5    & 1.5           & 5             \\
glass2                   & 5    & 5    & 5    & 5    & 5    & 5    & 5    & 5             & 5             \\
glass4                   & 8.5  & 8.5  & 5    & 7    & 3.5  & 1    & 3.5  & 2             & 6             \\
glass5                   & 7.5  & 7.5  & 4.5  & 9    & 1.5  & 4.5  & 6    & 3             & 1.5           \\
haberman                 & 3    & 3    & 3    & 3    & 7    & 9    & 7    & 3             & 7             \\
heart-stat               & 8    & 8    & 2    & 8    & 4.5  & 4.5  & 1    & 4.5           & 4.5      \vspace{1mm}\\ \hline \vspace{-3mm}     \\
\textbf{Average}         & 5.81 & 6.88 & 3.96 & 6.62 & 4.81 & 5.35 & 5.23 & \textbf{3.12} & \textbf{3.23}
\vspace{1mm}\\ \hline \vspace{-3mm} \\
\end{tabular}%
}
\end{table*}
\begin{table*}[]
\centering
\caption{Accuracy comparison with SOTA fuzzy baselines on UCI datasets.}
\label{tab:UCI_Accuracy}
\resizebox{\textwidth}{!}{%
\begin{tabular}{lcccccc} \hline \vspace{-3mm}\\  \textbf{\textbf{Dataset} $\downarrow$ $\mid$ \textbf{Model} $\rightarrow$}
 &
  NF-BLS \cite{feng2018fuzzy} &
  F-BLS \cite{sajid2024intuitionistic} &
  IF-TSVM \cite{rezvani2019intuitionistic} &
  edRVFL-FIS-C \cite{10552388} &
  IF-dRVFL (ours) &
  IF-edRVFL (ours)  \vspace{1mm}\\ \hline \vspace{-3mm} \\
acute\_nephritis              & 100   & 100   & 95    & 100   & 100            & 100            \\
blood                        & 77.57 & 75.7  & 76.24 & 76.77 & 76.37          & 76.37          \\
breast\_cancer                & 70.18 & 72.29 & 70.18 & 70.18 & 70.53          & 70.18          \\
echocardiogram               & 81.62 & 84.67 & 80.09 & 86.21 & 84.67          & 86.21          \\
haberman\_survival            & 73.49 & 69.66 & 73.49 & 73.49 & 73.49          & 73.49          \\
heart\_hungarian              & 68.99 & 78.94 & 64.41 & 71.36 & 76.87          & 78.89          \\
hepatitis                    & 88.39 & 84.52 & 79.35 & 86.45 & 87.1           & 87.87          \\
molec\_biol\_promoter          & 65.11 & 84.94 & 55.71 & 64.33 & 86.19          & 89.83          \\
pittsburg\_bridges\_T\_OR\_D & 89.14 & 88.19 & 86.14 & 89.14 & 86.14          & 86.14          \\
spect                        & 68.68 & 69.43 & 58.49 & 67.92 & 70.57          & 71.55          \\
tic\_tac\_toe                & 82.45 & 97.81 & 65.31 & 99.37 & 99.58          & 99.68          \\
titanic                      & 77.33 & 77.92 & 73.92 & 77.1  & 78.28          & 77.92          \vspace{1mm}\\ \hline \vspace{-3mm} \\
\textbf{Average}                      & 78.58 & 82    & 73.19 & 80.19 & \textbf{82.48} & \textbf{83.18}
\vspace{1mm}\\ \hline \vspace{-3mm} \\
\end{tabular}%
}
\end{table*}
\begin{table*}[]
\centering
\caption{Standard deviation comparison with SOTA fuzzy baselines on UCI datasets.}
\label{tab:UCI_Std}
\resizebox{\textwidth}{!}{%
\begin{tabular}{lcccccc} \hline \vspace{-3mm}\\  \textbf{\textbf{Dataset} $\downarrow$ $\mid$ \textbf{Model} $\rightarrow$}
 &
  NF-BLS \cite{feng2018fuzzy} &
  F-BLS \cite{sajid2024intuitionistic} &
  IF-TSVM \cite{rezvani2019intuitionistic} &
  edRVFL-FIS-C \cite{10552388} &
  IF-dRVFL (ours) &
  IF-edRVFL (ours)  \vspace{1mm}\\ \hline \vspace{-3mm} \\
acute\_nephritis              & 0     & 0     & 49.35 & 0     & 0              & 0              \\
blood                        & 12.26 & 11.45 & 14.99 & 13.44 & 15.15          & 15.15          \\
breast\_cancer                & 44.62 & 27.59 & 44.62 & 44.62 & 44.44          & 44.62          \\
echocardiogram               & 4.39  & 7.27  & 7.01  & 7.05  & 7.76           & 7.05           \\
haberman\_survival            & 8.48  & 10.07 & 8.48  & 8.48  & 8.48           & 8.48           \\
heart\_hungarian              & 14.98 & 9.58  & 48.5  & 15.47 & 10.36          & 14.31          \\
hepatitis                    & 8.72  & 9.51  & 14.17 & 8.66  & 9.12           & 7.57           \\
molec\_biol\_promoter          & 12.79 & 3.79  & 26.21 & 16.61 & 13.72          & 5.27           \\
pittsburg\_bridges\_T\_OR\_D & 9     & 8.36  & 13.95 & 9.67  & 13.95          & 13.95          \\
spect                        & 7.38  & 2.8   & 14.06 & 4.81  & 3.43           & 14.02          \\
tic\_tac\_toe                & 14.76 & 1.7   & 48.42 & 1.14  & 0.94           & 0.93           \\
titanic                      & 13.96 & 15.58 & 15.91 & 15.93 & 13.78          & 13.73          \vspace{1mm}\\ \hline \vspace{-3mm} \\
\textbf{Average}                      & 12.61 & 8.98  & 25.47 & 12.16 & \textbf{11.76} & \textbf{12.09}
\vspace{1mm}\\ \hline \vspace{-3mm} \\
\end{tabular}%
}
\end{table*}
\begin{table*}[]
\centering
\caption{Rank comparison with SOTA fuzzy baselines on UCI datasets.}
\label{tab:UCI_Rank}
\resizebox{\textwidth}{!}{%
\begin{tabular}{lcccccc} \hline \vspace{-3mm}\\  \textbf{\textbf{Dataset} $\downarrow$ $\mid$ \textbf{Model} $\rightarrow$}
 &
  NF-BLS \cite{feng2018fuzzy} &
  F-BLS \cite{sajid2024intuitionistic} &
  IF-TSVM \cite{rezvani2019intuitionistic} &
  edRVFL-FIS-C \cite{10552388} &
  IF-dRVFL (ours) &
  IF-edRVFL (ours)  \vspace{1mm}\\ \hline \vspace{-3mm} \\
acute\_nephritis              & 3    & 3    & 6    & 3    & 3             & 3             \\
blood                        & 1    & 6    & 5    & 2    & 3.5           & 3.5           \\
breast\_cancer                & 4.5  & 1    & 4.5  & 4.5  & 2             & 4.5           \\
echocardiogram               & 5    & 3.5  & 6    & 1.5  & 3.5           & 1.5           \\
haberman\_survival            & 3    & 6    & 3    & 3    & 3             & 3             \\
heart\_hungarian              & 5    & 1    & 6    & 4    & 3             & 2             \\
hepatitis                    & 1    & 5    & 6    & 4    & 3             & 2             \\
molec\_biol\_promoter          & 4    & 3    & 6    & 5    & 2             & 1             \\
pittsburg\_bridges\_T\_OR\_D & 1.5  & 3    & 5    & 1.5  & 5             & 5             \\
spect                        & 4    & 3    & 6    & 5    & 2             & 1             \\
tic\_tac\_toe                & 5    & 4    & 6    & 3    & 2             & 1             \\
titanic                      & 4    & 2.5  & 6    & 5    & 1             & 2             \vspace{1mm}\\ \hline \vspace{-3mm} \\
\textbf{Average}                      & 3.42 & 3.42 & 5.46 & 3.46 & \textbf{2.75} & \textbf{2.46}
\vspace{1mm}\\ \hline \vspace{-3mm} \\
\end{tabular}%
}
\end{table*}

\newpage
\bibliographystyle{splncs04}
\bibliography{refs.bib}